\documentclass[format=sigconf]{acmart} 

\usepackage{multicol,caption} 
\usepackage{enumitem} 
\usepackage{mdframed} 
\usepackage{amsmath} 
\usepackage{titlesec} 
\usepackage{tabularx}
\usepackage[detect-all]{siunitx} 

\usepackage{balance} 

\usepackage{caption}
\usepackage{subcaption}

\newtheorem{mydef}{Definition}
\AfterEndEnvironment{mydef}{\noindent\ignorespaces}

\newenvironment{myalg}[1]
  {\mdfsetup{
    frametitle={\colorbox{white}{\space#1\space}},
    innertopmargin=0pt,
    frametitleaboveskip=-\ht\strutbox,
    frametitlealignment=\center
    }
  \begin{mdframed}%
  }
  {\end{mdframed}}

\def\eps{\pmb{\varepsilon'}}

\def\and{{\rm and}}

\titleformat*{\section}{\LARGE\bfseries}
\titleformat*{\subsection}{\Large\bfseries}
\titleformat*{\subsubsection}{\large\bfseries}
\titleformat*{\paragraph}{\large\bfseries}
\titleformat*{\subparagraph}{\large\bfseries}
\theoremstyle{plain}

\titlespacing*{\section}
{0pt}{1.9ex plus 0.7ex minus .2ex}{0.8ex plus .2ex}

\titlespacing*{\subsection}
{0pt}{1.2ex plus 0.8ex minus .2ex}{0.7ex plus .2ex}

\titlespacing*{\subsubsection}
{0pt}{0.4ex plus 0.1ex minus .2ex}{0.5ex plus .2ex}

\makeatletter
\newsavebox{\mybox}\newsavebox{\mysim}
\newcommand{\distras}[1]{%
  \savebox{\mybox}{\hbox{\kern3pt$\scriptstyle#1$\kern3pt}}%
  \savebox{\mysim}{\hbox{$\sim$}}%
  \mathbin{\overset{#1}{\kern\z@\resizebox{\wd\mybox}{\ht\mysim}{$\sim$}}}%
}
\makeatother

\begin{document}

\graphicspath{{figures/}}

\title{Intrinsic Point of Interest Discovery from Trajectory Data}


\author{Matthew Piekenbrock}
\affiliation{
Dept. of Computer Science \& Engineering \\ 
Kno.e.sis Research Center \\
Wright State University, Dayton, OH, USA \\
piekenbrock.5@wright.edu}

\author{Derek Doran}
\affiliation{
Dept. of Computer Science \& Engineering \\
Kno.e.sis Research Center \\
Wright State University, Dayton, OH, USA \\
derek.doran@wright.edu}

\date{}

\begin{abstract}
This paper presents a framework for intrinsic point of interest discovery from trajectory databases. 
Intrinsic points of interest are regions of a geospatial area innately defined by the spatial and temporal aspects of trajectory data, and can be of varying size, shape, and resolution. 
Any trajectory database exhibits such points of interest, and hence are {\em intrinsic}, as compared to most other point of interest definitions which are said to be {\em extrinsic}, as they require 
trajectory metadata, external knowledge about the region the trajectories are observed, 
or other application-specific information. 
Spatial and temporal aspects are qualities of any  trajectory database, making the framework
 applicable to data from any domain and of any resolution.
The framework is developed under recent developments on the consistency of nonparametric hierarchical density estimators and enables 
the possibility of formal statistical inference and evaluation over such intrinsic points of interest. 
Comparisons of the POIs uncovered by the framework in synthetic truth data to thousands of parameter settings for common
POI discovery methods show a marked improvement in fidelity
without the need to tune any parameters by hand.
\end{abstract}
\maketitle

\section{Introduction}\label{sec:introduction}
The development and deployment of location acquisition systems have enabled large scale capturing of `movement' or `trajectory' data from people, cars, and other objects. Technologies like global positioning systems (GPS), global system for mobile communications (GSM), wide area motion imagery (WAMI), and radio-frequency identification (RFID) allow organizations and governments to collect and exploit trajectory patterns in many scenarios.
More recent initiatives (e.g. Uber's Movement\footnote{\url{https://movement.uber.com/cities}} and IBM's Smarter Cities\footnote{\url{https://www.ibm.com/smarterplanet/us/en/smarter_cities/overview/}} programs) have even made such data available to the either the public or city planning experts at large. 
With the rise in importance of this data comes prevalent use of Geographic Information Systems (GIS) and related platforms such as ArcGIS\footnote{\url{https://www.arcgis.com/}} and Mapbox\footnote{\url{https://www.mapbox.com/}}. Other related use cases of GIS information have also emerged for surveillance~\cite{chang2009combining} and location-based service (LBS) applications~\cite{virrantaus2001developing}. In many of these applications, trajectory data is exploited for knowledge acquisition tasks~\cite{gonzalez2008understanding}, the integration of movement patterns to uncover ``patterns of life" over a region~\cite{zheng2010geolife}, to expand situational awareness in crises~\cite{zhang2017emergency}, and to support the value added by a LBS application~\cite{figueiredo2016tribeflow}.

In many of these knowledge acquisition tasks, the notion of a ``location" or ``point of interest" (POI) is foundational to understanding the entirety of the common space in which the data are observed~\cite{pavan2015finding}. For example, mapping systems must know the position and geometry of locations for navigation and automated guidance control purposes. In LBS applications, the POIs and metadata such as their popularity (e.g. `star-rating') are necessary to provide useful location recommendations~\cite{zheng09,park2007location, figueiredo2016tribeflow}. 
Because POIs are not available from `raw' trajectory data captured by location acquisition systems, they are often {\em extrinsically} defined by gazetteers such as Google Places, FourSquare, GeoNames, or OpenStreetMap. 	
Yet external sources of location data present many difficulties when faced with the problem of understanding how a given trajectory dataset relates to the underlying geographical area where it was observed. For example, many gazetteers store varying types of either POI metadata or POI relational data, allowing gazetteer-derived information to present a source of {\em bias}. Furthermore, relying on gazetteers explicitly {\em defines} the set of POIs that exist in a given geographical region. When there  is disagreement on this definition, analysis becomes difficult. Furthermore, with POIs defined {\it a priori}, one is faced with the problem of ``fitting'' observed trajectory data to models defined by such POIs, many of which may or may not be relevant to the given data at hand. For example, it may be desirable for a city-planner gathering movement (trajectory) data following a public event to discover `bottleneck' congestion areas like parking lots, roads, or sidewalk segments for the purpose of traffic analysis. In this situation, it would more useful to discover POIs directly {\em from the data itself} during the event, but such geographical POIs may not be available in a gazetteer. 

To over come these challenges, this paper investigates the POI discovery problem in the most generic context possible. We ask: {\em given only trajectory data, without access to gazetteers, can we infer subregions within a geospace that are ``interesting" enough to call it a POI?}
We seek {\em intrinsic} POIs, which are POIs recoverable without the use of a gazetteer, are completely defined by observed movement patterns, and can be used for any domain-specific application and at any scale (e.g. from movements within a building to movements across an entire city region). 
To make such a definition meaningful, we build off recent theoretical work in density-based clustering and introduce a data-driven, statistically rigorous definition of a POI applicable to trajectory data of any (and even mixed) resolution. The definition follows from a recent minimax analysis of the consistency of hierarchical density superlevel set estimators. We use this definition to present a {\bf parameter-free} framework for extracting intrinsic POIs, i.e. yields an optimal unsupervised 
solution without ad hoc parameter-tuning.\footnote{We see this as a necessary form of {\bf usability}, an important feature to have in the modern clustering era. It is well-known that having several sensitive, real-valued parameters results in combinatorial explosion of the parameter space of an algorithm, resulting in the need for the user to use one or more parameter-tuning methods to arrive at a solution that befits the application.}
A comparative analysis is performed on realistic simulations involving both vehicle and pedestrian traffic. Validation results show marked improvements in fidelity against several state of the art (SOTA) algorithms. Of interest to the authors, the simulation settings, the resulting traffic data, the validation code, and the framework itself is all {\bf completely reproducible} and {\bf open source}, available online.\footnote{<Anonymized for review purposes.>}
\section{Point of Interest Discovery}\label{sec:methodology}

This section provides preliminary information about the POI discovery problem, and provides context and definitions for this work. It then formally defines a POI and (subsequently) an intrinsic POI, and the framework their discovery. 
\subsection{Preliminaries} 
We consider a trajectory database of discrete, time-indexed spatial data having at least the 3-tuple of attributes
$$(\texttt{<object id>}, \texttt{<spatial component>}, \texttt{<temporal component>})$$ 
This minimal amount of information implies a {\it trajectory} for an object of the form: 
\begin{equation}\label{eq:1}
	T = p_1 \xrightarrow[]{\Delta t_1} p_2 \xrightarrow[]{\Delta t_2} \dots \xrightarrow[]{\Delta t_{n-1}} p_n
\end{equation}
where $p_1, p_2, ..., p_n$ are chronologically ordered spatial coordinates. In the geographical sense, these spatial components are often defined by a \texttt{<latitude>} and \texttt{<longitude>} pair, but in practice could be from any coordinate system. 
Such representations require {\it trajectory pattern mining} techniques~\cite{zheng2015trajectory}, or techniques that seek to mine common spatiotemporal patterns across trajectories to assert significance over areas where trajectory patterns emerge. Mined patterns in trajectories are often referred to as {\it mobility patterns}, characterizing some specific trajectory quality of interest, such as heading, stopping rate, velocity, rotation, curvature, or shape~\cite{buchin2011segmenting}. Such mobility patterns exhibit properties that make the formal retrieval of significant areas challenging. For example, if the timespan of an observed trajectory is long, the processes driving the mobility pattern may be non-stationary (e.g. road traffic that changes due to construction, or congestion effects due to time of day shifts in the work schedule). There may also be paths of objects that are transient (some areas are never traveled to more than once). Furthermore, the spatial components in trajectory data often have a high degree of autocorrelation, breaking assumptions of independence~\cite{figueiredo2016tribeflow}. A variety of models have been proposed to handle thee situations, largely focusing on estimating {\it individual} trajectory statistics under these assumptions. This includes, for examples, adaptive Kalman filters for vehicle navigation~\cite{hu2009adaptive}, state-space models~\cite{gindele2010probabilistic}, and trajectory path uncertainty models~\cite{trajcevski2009continuous}.

The knowledge mined from individual trajectories says little of the {\it macroscopic patterns driving such trajectory observations}. Rather than focusing on the statistics of individual trajectories,
{\em collective models} preprocess the trajectory data to extract characteristics across a swath of trajectories. Such preprocessing is desirable, as it discards highly autocorrelated data representing redundant information in favor of aggregating trajectory positions into observations of significance. Examples of this preprocessing scheme include extracting ``semantically enriched'' points that intersect known geographical regions~\cite{alvares2007model}, aggregating trajectory positions as stay points using supplied spatial and/or temporal thresholds~\cite{zheng2010geolife}, or processing trajectory data into groups using some  convex combination of spatial, temporal, and semantic similarity kernels~\cite{liu2013todmis, ying2011semantic}. 

In a collective model, we refer to the `important' or semantically meaningful data samples aggregated from trajectory points as {\it exemplar positions}, or simply, {\it exemplars}:
\begin{mydef} {\bf Exemplar}\\
	Consider a sample of $n$ discrete points $X_n \subset \mathbb{R}^d$ that 
	constitute a trajectory $T$. 
	Define an aggregation function $\alpha : \mathcal{P}(X_n) \mapsto \mathbb{R}^d$ 
	that maps any subset of points (e.g. a trajectory segment) in $T$ to a set of {\bf exemplar} positions $\Sigma \subset \mathbb{R}^d$. 
\end{mydef}
\noindent
The aggregation function of choice depends on the intent of the analysis. For example, consider an urban environmental study that defines $\alpha$ as a mapping of some isolated trajectory {\it segment} $\{p_k, p_{k+1}, ..., p_{k+l}\}$ to the mean coordinate of the segment if the speed of the object traveling from $p_k$ to $p_{k+l}$ exceeds a certain threshold. Groups of these exemplar positions may determine ``high-emission'' zones in a city~\cite{baluja2016reducing}. Alternatively, if the traffic is made of pedestrians, such groups may represent tourist attraction areas, the popularity of which are useful for LBS applications~\cite{zheng09}. 
It is not difficult to find this type of trajectory preprocessing in geospatial applications, and the grouping of them is foundational to countless tasks in trajectory mining~\cite{zheng2010geolife, alvares2007model, liu2012tra, zhou07, zheng09, ying2011semantic}. We generalize this preprocessing step by referring to it as ``exemplar extraction.''

An important aspect of exemplar extraction is to choose an aggregation function that befits the intent of the analyst and thus satisfies a study's interpretation of ``interesting.'' This is inevitably application-specific, and the proposed framework is agnostic to the specific form of aggregation used, thus it is irrelevant to bestow a particular interpretation of what ``interesting" means. We consider a more concrete and practical definition using a popular type of aggregation in Section \ref{sec:experiments_and_discussion}. 

\begin{figure*}
 	\includegraphics[width=\textwidth, height=5.95cm]{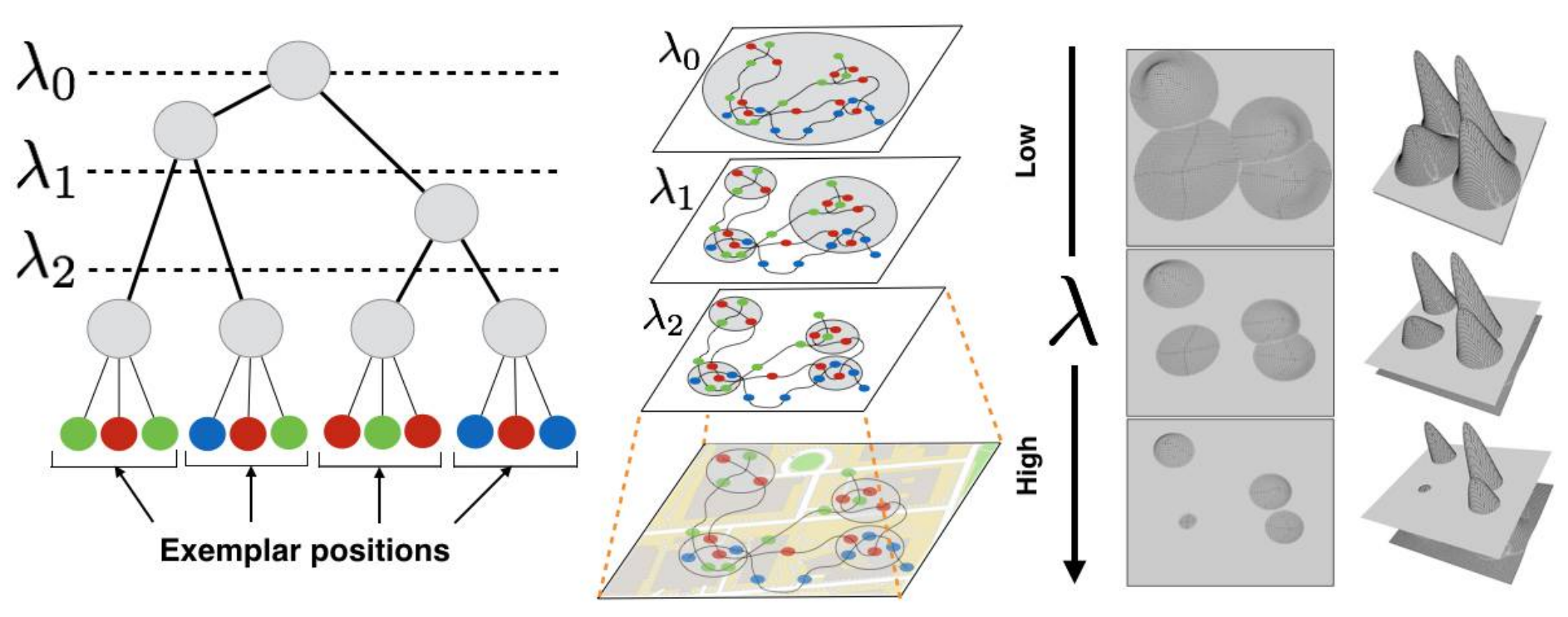}
 	\caption{Illustrating the cluster tree hierarchy and its interpretation of POIs. Consider an estimated density (right panel) of exemplar positions extracted from trajectories in a geospace (middle, bottom panel). A POI is a geospatial region inhabited by exemplar positions at some density threshold $\lambda$, with the number of the POIs extracted depend on this scale parameter setting (left panel). Higher $\lambda$ limits a POI to being specific and small, and could cause POIs to be manifested by random noise or be overfitted to a particular set of observations. Low $\lambda$ defines POIs as very broad areas of low exemplar position density. The cluster tree hierarchy (left panel) summarizes the set of exemplar positions representing a POI at every density threshold, thus capturing the entire collection of POIs over a common area (middle panel, upper layers). }
 	\label{fig:ctree}
\end{figure*}

\subsection{Defining a point of interest}\label{sec:definition}
Under the premise that exemplars represent meaningful aggregations of observations from a trajectory data source, it is natural to define a POI as {\em a region of exemplars}. We seek a definition of such regions with a statistical (rather than heuristic) foundation as a means of reflecting the naturally occurring structure within the data. Towards this end, we define a POI as a {\em contiguous, high density region of exemplars}. 

To formalize this definition, we follow the notation of Chaudhuri et. al~\cite{chaudhuri2010rates}. Let $\mathcal{X}$ be a subset of $\mathbb{R}^d$ and define a {\em path} as a function $P : [0,1] \to S$ where $S \subset \mathcal{X}$. Also denote the equivalence relation $C$ as {\em connected}, where $xCy$ iff $P(0) = x$ and $P(1) = y$. Then $C$ partitions $S$ into {\it connected components} or {\em clusters}. Each component represents an area of high density and is called a {\em high density cluster}:
\begin{mydef} {\bf High density clusters}\\
For a density function $f$ on $\mathbb{R}^d$, consider a partitioning:
	\begin{equation}\label{eq:poi}
		\{ x : f(x) \geq \lambda \}\text{, for some }\lambda > 0
	\end{equation}
where $\lambda$ is called the level, or high density threshold, parameter. Then all maximally connected components in this set are {\bf high density clusters} at density level $\lambda$. 
\end{mydef}
We relate this formal definition to the trajectory mining domain with the following definition of a point of interest, defined over {\em a extracted set of exemplars}. 
\begin{mydef} {\bf Point of interest}\\
Given a set of $m$ exemplars $\{\varepsilon_1, \varepsilon_2, \dots, \varepsilon_m\} \in \Sigma$ and a fixed ``scale'' or resolution $\lambda$, each high density cluster of such exemplars forms a {\bf point of interest} at the density level $\lambda$. 
\end{mydef}
The sets of high density clusters across all values of $\lambda$ forms a hierarchy often referred to as the {\bf cluster tree} of the density $f$~\cite{chaudhuri2010rates, chen2016statistical}. 
A hierarchical definition of locations is common~\cite{zheng09} and matches the intuitive interpretation of a POI. For example, not only may a particular restaurant in a mall food court be a POI, but the food court itself may also be considered a POI, as well as entire mall may be yet another POI. The cluster tree conceptualization formalizes a POI as a maximally connected set of exemplars falling along a higher density area, implying such areas are `significant,' and that such connected exemplars may be related.    

A visualization of hierarchical POIs and a dendrogram of the corresponding cluster tree is provided in Figure~\ref{fig:ctree}. The middle figure demonstrates a high-level view of what a set of trajectories might look like, with the colored dots in the left and middle figures representing exemplars. The right figure demonstrates a density estimate of the positions of these exemplars. That is, when these exemplars are very close to each other, they're said {\em to have a high density and are thought to be related}, constituting a POI---the scale of the density depends on the analysis at hand. A sufficiently low density threshold $\lambda_0$ will designate every exemplar as one POI.

From this definition, it may seem that any arbitrary density estimator may be used to find high-density clusters: simply estimate the density of every point by kernel density estimation (KDE), and then iterate through all possible values of $\lambda$ that create distinct high-density clusters. Yet not every estimation will produce the same hierarchy---different kernels (and kernel bandwidths) may result in a completely different hierarchy of high-density clusters, and by extension, a different set of POIs.  
From the cluster tree perspective, the ideal kernel $f_n$ is one that is uniformly consistent (i.e. $\sup_x{\lvert f_n(x) - f(x) \rvert} \to 0 \text{ as } n \to \infty$) from a given sample $X_n$. In this case,  a model could be fitted with the appropriate kernel and bandwidth parameter, and the would KDE furnish a continuous surface from which a cluster tree and its high-density clusters can be derived~\cite{chen2016statistical}. The main issue is that the set of all high-density clusters is not easy to compute for typical density estimates of $f$~\cite{chaudhuri2010rates} and generally require a significant amount of memory to store. This computational inefficiency limits usability for large trajectory datasets, often observed over wide geographical areas and over long periods of time. 
From the applied perspective, many state-of-the-art approaches find POIs by variants of hierarchical clustering to find groups of exemplars. This has proved useful for application-specific problems~\cite{zheng2010geolife, zhou07, zheng09} but they are largely heuristic, i.e. it is common for most clustering algorithms to have unstated or unknown statistical properties, precluding the possibility of formal inference~\cite{fraley2002model}. The framework we introduce therefore examines
{\em density-based} clustering methods as they are 
designed to infer a cluster tree~\cite{chen2016statistical}
without facing the computational hurdles of KDEs. 

A desirable property of any finite-sample density estimator is some notion of {\em consistency}.\footnote{Recall that an estimator $\hat{\Theta}_n$ whose value $\hat{\theta}$ is a point estimate of $\theta$ is consistent if, as more samples are collected ($n \to \infty$), $\hat{\Theta}_n$ {\bf converges in probability} to the true value of the parameter, i.e. $\underset{n \to \infty}{\mathrm{plim}}(\hat{\Theta}_n) = \theta$.} In 1981, Hartigan establsihed a reasonable definition~\cite{hartigan1981consistency}, often referred to as {\it Hartigan consistency}:
\begin{mydef}{\bf Hartigan Consistency}\\
Let $\mathbb{C}_{f_n}$ be the set of all high-density clusters from the cluster tree. 
For any sets $A, A' \subset \mathcal{X}$, let $A_n$ (respectively, $A_n'$) denote the smallest set of $\mathbb{C}_{f_n}$ containing $A \cap X_n$  (respectively $A' \cap X_n$). $\mathbb{C}_{f_n}$ is consistent if, whenever $A$ and $A'$ are different connected components of $\{ x : f(x) \geq \lambda \}$ (for some $\lambda > 0$), $\mathbb{P}(A_n \text{ is disjoint from } A_n') \to 1$ as $n \to \infty$.
\end{mydef}
This consistency definition essentially requires that two disjoint high-density clusters from the unknown, population density ($A$ and $A'$) will also be disjoint components in a given empirical cluster tree ($A_n$ and $A_n'$), given enough samples ($n$). 
The proposed framework for POI discovery is developed and implemented from the first computationally tractable and provably consistent algorithm that satisfies Hartigan consistency, as analyzed by Chaudhuri {\em et. al}~\cite{chaudhuri2010rates}, to be discussed in the next section. 
Having a nonparametric model satisfying this notion of consistency is important, as it transforms the unsupervised problem of POI discovery into a formal statistical estimation problem, not only enabling analysis {\it driven by data}, but requiring minimal assumptions regarding the nature of the data. 
Such a relation enables methods of formal statistical inference, allowing one to quantify uncertainty, i.e. to create hypothesis tests to discern ``true'' POIs as opposed to ``false'' POIs resulting from random noise or artifacts of low sample sizes, or to create notions of confidence in estimation~\cite{chen2016statistical}.

\subsubsection*{Consistent cluster tree estimation:}
We next motivate a recent cluster tree estimator and discuss its relationship and applicability to POI discovery for 
the propsoed framework. Recall that an empirical estimate of the cluster tree, applied over exemplars, represents a hierarchy of POIs. Viewed from this perspective, what we propose can be seen as an extension of Chaudhuri {\em et. al}'s work on the cluster tree~\cite{chaudhuri2010rates} to a trajectory mining context. 

\begin{figure}
		\includegraphics[width=0.34\textwidth, height=3.6cm]{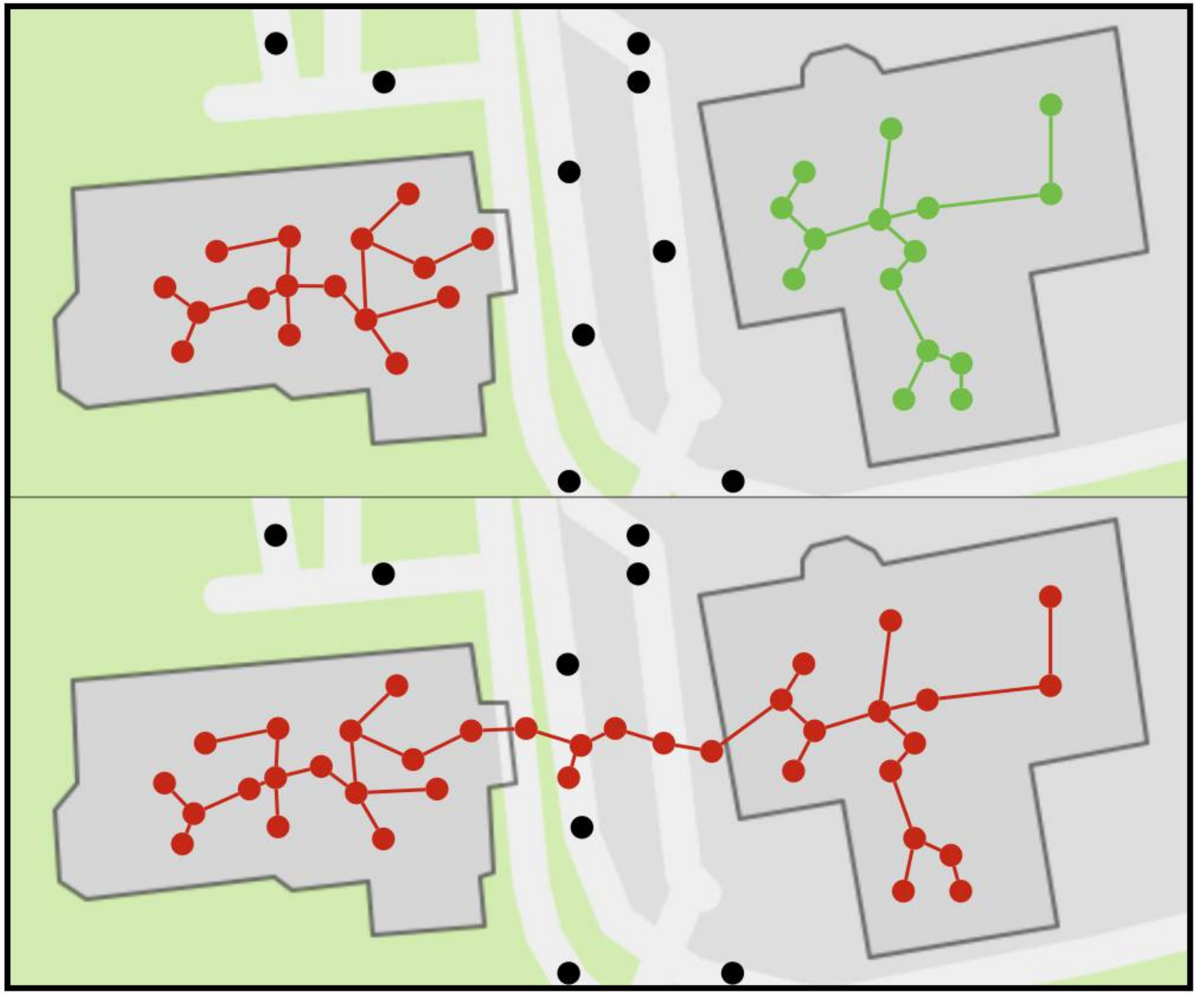}
		\caption{SL excessive chaining example. The bottom panel denotes a possible clustering using SL when pedestrians were found to stop between buildings. }\label{fig:sl_chaining}
\end{figure}

Consider using Single-Linkage (SL) clustering, an agglomerative scheme that creates a hierarchical representation of clusters using the minimum pairwise distance $D$ between all points, as a tool for clustering exemplars. Beginning with every exemplar $x$ as a singleton, SL iteratively merges exemplars into clusters according to the {\it linkage function}: 
$$ D(x_i, x_j) = \min\limits_{x_i, x_j \in X} d(x_i, x_j) $$
SL clustering is often criticized due to its tendency to create `excessive chaining', wherein two clusters which may have been seen as generally unrelated are amalgamated by chance at a distance threshold that does not reflect the true dissimilarity between the resulting clusters. Hartigan proved SL is a consistent estimator of the cluster tree for densities in $\mathbb{R}$ (for $d = 1$) and is {\em not consistent} for any $d > 1$~\cite{hartigan1981consistency}, implying that any SL cluster that contains all the sample points in $A$ will also contain nearly all sample points in $A'$, in probability as $n \to \infty$.\footnote{The condition is related to the ``thin bridge'' between any two population modes. Fractional consistency was shown for SL if for any pair \unexpanded{$A$} and \unexpanded{$A'$}, the ratio of \unexpanded{$\inf \{ f(x) : x \in A \cup A' \}$ to $\sup \{ \inf \{ f(x) : x \in P \} : \text{paths } P \text{ from } A \text{ to } A' \}$} is sufficiently large.}
This is reflected in the geospatial sense as well: consider the case where exemplars represent aggregated `stops' within a set of trajectories, a case that we will also consider later in Section \ref{sec:experiments_and_discussion}. If an area is observed long enough, such exemplars should naturally form an area high density in areas where people stop frequently, e.g. within buildings. In such cases, it may be useful to categorize exemplars within their respective POIs (this is done in supervised way applications extract semantic information, see ~\cite{alvares2007model} for example). However, it's also possible that there exist a few stops just outside of such buildings, which SL has a tendency to chain together. An example of this is shown in Figure \ref{fig:sl_chaining}.
This discovery motivated efforts to modify SL not only to reduce this chaining to make SL more `robust', but also to achieve (at least) Hartigan consistency for $d = 2$ and beyond. The first provably consistent estimator, which we consider in this effort, is a generalization of SL referred to as `Robust Single Linkage' (RSL)~\cite{chaudhuri2010rates}. 

{\bf Robust Single Linkage:} Let $\mathcal{X}$ be a subset of $\mathbb{R}^d$. Let $\lVert \cdot \rVert$ denote the $\ell_2$ norm and let $B(x, r)$ be a closed ball of radius $r$ around the point $x$. 
The RSL algorithm is given in the listing below: 
\vbox{%
\begin{myalg}{Robust Single Linkage Algorithm}\label{alg:rsl}
\begin{enumerate}[leftmargin=*]
  \item For each $x_i$ set $r_k (x_i) = \mathrm{inf}\{ r : B(x_i, r)$ contains $k$ data points$\}$.
  \item As $r$ grows from 0 to $\infty$:
  \begin{enumerate}
    \item Construct a graph $G_r$ with nodes $\{x_i : r_k(x_i) \leq r \}$. \\
          Include edge $(x_i, x_j)$ if $\lVert x_i - x_j \rVert \leq \alpha r$
    \item Let $\mathbb{C}_{f_n}(r)$ be the connected components of $G_r$.
  \end{enumerate}
\end{enumerate}
\end{myalg}
}
The RSL algorithm has two free parameters which need to be set: $\alpha$ and $k$. SL is equivalent to RSL with the setting $\alpha = 1$, $k = 2$. 
Whereas SL is equivalent to (and can be efficiently computed by) the minimum spanning tree (MST) computed over all pairwise distances, RSL scales these distances by a constant factor $\alpha$, and only reduces to the MST if the components are restricted from connecting (satisfying $\{x_i : r_k(x_i) \leq r \}$) {\em within the MST computation}. Chaudhuri {\em et. al} found that RSL is Hartigan consistent and established finite-sample rates of convergence for all $1 \leq \alpha \leq 2$, with the optimal rate of convergence with the setting $\alpha \geq 2$~\cite{chaudhuri2010rates}.

 \subsection{Finding intrinsic points of interest}
Using a consistent cluster tree estimator, such as RSL, on a set of exemplars creates a hierarchical representation of POIs. However, a nested set of multiple solutions is not always desirable, and a `flat' solution (where each point is assigned a single label) may be preferred. A traditional approach in hierarchical clustering, ``cutting'' the empirical cluster tree at a given density threshold value $\lambda$ yields a set of high-density clusters
$\mathbb{C}_{f_n}(\lambda) = \{C_1, C_2, \dots C_m\}$
that form $m$ POIs, a possible `flat' solution. 
However, the choice of $\lambda$ forces all POIs to be of the same scale, requires the user to know which  granularity to choose {\em a priori}, affecting the size and kinds of POIs discovered. For example, a small $\lambda$ may define shops in a mall as POIs, while a larger $\lambda$ may define the mall itself as a POI. It may not be known ahead of time what granularity level is relevant. 
Furthermore, it is reasonable to expect that relevant POIs exist at multiple levels of granularity, such that a sprawling city park and a small restaurant could both constitute a POI. 
Thus, it would useful to have some sensible notion of ``cluster quality'' that can be used (and optimized) as an objective function to discover POIs that are not dependent on the analyst's choice of $\lambda$, and are {\em strongly intrinsic to the geospace itself}, e.g. are {\bf intrinsic POIs}. 

To capture POIs of any scale and hence satisfy our notion of an intrinsic POIs, we first recall that high-density clusters are contiguous, relatively dense areas of the data space, separated by contiguous and relatively non-dense areas, defined over a working definition of density over a set of exemplars. From a statistical point of view, we can think of a high-density cluster as a set of points with high density around some ``neighborhood'' or volume of the support. M{\"u}ller {\em et. al} quantify this using a functional called {\em excess of mass}~\cite{muller1991excess}: 
\begin{mydef}  {\bf Excess of Mass}\\
For a $C_i \in \mathbb{C}_{f_n}(\lambda)$ for some value of $\lambda > 0$, the {\bf excess of mass} of $C_i$ is given by: 
	\begin{equation}\label{eq:4}
		E(C_i) = \int_{x \in C_i} \Big( f(x) - \lambda_{min}(C_i) \Big)\text{ }dx
	\end{equation}
\end{mydef}
where $\lambda_{min}(C_i)$ represents the lowest density level $C_i$appears.
Initially, this measure seems like a reasonable definition of the ``quality" of a clustering within the cluster tree estimate. Considering the definition of a high-density cluster from Equation \ref{eq:poi}, where a cluster exists along a mode of local maximum of the underlying density, it's far too likely that a finite-sample estimation may empirically find a mode at a given point $x_0$ if the data is sparse, allowing an arbitrarily low probability associated $x_0$ to be classified. A more interesting result would be to associate a high-density cluster with a region that exhibits relatively high probability {\em over a  neighborhood}. See M{\"u}ller {\em et. al} for visualization, along with a more in depth description of this functional~\cite{muller1991excess}. 
However, as Campello {\it et. al} remark, this measure exhibits monotonic behavior in any direction varying the density-level $\lambda$ in the hierarchy, and instead propose an alternative, {\em local measure} of cluster quality~\cite{campello2015hierarchical}:
\begin{mydef}  {\bf Relative Excess of Mass}\\
For a $C_i \in \mathbb{C}_{f_n}(\lambda)$ for some value of $\lambda > 0$, the {\bf relative excess of mass} of $C_i$ is given by: 
	\begin{equation}\label{eq:4}
		E_R(C_i) = \int_{x \in C_i} \Big( \lambda_{max}(x, C_i) - \lambda_{min}(C_i) \Big)\text{ }dx
	\end{equation}
\end{mydef}
where $\lambda_{max}(x, C_i) = \min\{ f(x), \lambda_{max}(C_i)\}$ is the density level beyond which $x$ is no longer part of $C_i$, and $\lambda_{max}(C_i)$ is the highest density beyond which $C_i$ either becomes disconnected (creating separate components) or disappears (creating singleton clusters). It is important to note that relative excess of mass is defined in terms of $\lambda$ values associated with a specific {\em cluster}, as opposed to a specific {\em clustering}. 
This implies that an `optimal' clustering with respect to the relative excess of mass estimate may not occur at a fixed,  {\em global} density threshold, but rather as a result of several {\em local} density thresholds applied to the hierarchy. Intuitively, if a given cluster $C_i$ contains many points that have high density relative to $\lambda_{min}(C_i)$, such a cluster will exist across several thresholds of $\lambda$ and is thus robust to fluctuations in the scale of analysis. For this reason, the relative excess of mass can be thought of as a measure of cluster `stability' across different density levels, which we posit reflects an {\em intrinsic} POI that is innately defined by the dataset independent of density level. Such intrinsic POIs are thus defined as follows: let $\delta_i$ be an indicator equal to 1 if cluster $C_i \in \mathbb{C}_{f_n}$ represents an intrinsic POI and 0 otherwise. Assign values to these indicators such that the following is maximized: 
 \vspace*{-0.085cm}
 \begin{equation}\label{eq:10}
 	\begin{aligned}
 		& \underset{\delta_1, \dots, \delta_m}{\text{maximize}}
 		& & J = \sum\limits_{i = 2}^m \delta_i E_R(C_i) \\
 		& \text{subject to}
 		& & 
 		\begin{cases} \delta_i \in \{ 0, 1 \}, i \in \{2, \dots, m\} \\
 		\text{exactly one }\delta(\cdot) = 1 \text{ per disjoint branch}
 		\end{cases}
 		\end{aligned}
 \end{equation}
 Where the ``per disjoint branch'' constraint means that the indicator function $\delta(\cdot)$ equals 1 exactly once for all clusters in each path from a leaf node to the root of the cluster tree. The optimization of this objective function is beyond the scope of this paper; we refer to Campello {\em et. al}'s cluster extraction method for general cluster hierarchies~\cite{campello2013framework} to solve this optimization, as it was developed alongside an estimator very similar to RSL, is capable of producing an optimal result at several density levels, and accounts for the density thresholds at which points become noise (fall along densities below a given threshold). 

\section{Experiments and Discussion}\label{sec:experiments_and_discussion} 
We next evaluate the proposed framework for intrinsic POI discovery. Because intrinsic POIs do not rely on gazetteers and may manifest themselves in unknown locations, evaluation on real data validated against ``ground truth" external knowledge (such as imported location data from sources such as OpenStreetMap or Google Places) is not feasible. A common approach to evaluate clusterings when ground truth is absent is to use an internal cluster validity index (CVI). 
CVIs include common indices like the Silhouette score, the Dunn Index, and the Calinski-Harabasz criterion (see Arbelaitz ~\cite{arbelaitz2013extensive} for an overview of these techniques and references therein). Recent work recommends validation using multiple CVIs, as they each score different aspects of a clustering such as the ratio of inter- to intra-cluster distances, sum of squares distance to centroid, or graph-theory scores based on similarity~\cite{arbelaitz2013extensive}. 
We do not believe scores are informative for intrinsic POI evaluation, as most of these CVIs operate on unrealistic assumptions (e.g. symmetry or convexity of cluster shape, a notion of minimal variance, the existence of a centroid or medoid, etc.). Contrary to these widespread concepts, we do not assume that a cluster of exemplars representing an intrinsic POI will maintain some hyper-spherical or -elliptical shape. 
Indeed, there are a number of features within a geographical area that may be considered POIs, yet inevitably exhibit arbitrary shapes (e.g. buildings, parks, gathering areas, etc.) and manifest at varying densities (e.g. a busy intersection that is small and concentrated in exemplar density vs. a parking lot that is large and more uniform in density). 
Following the advice of Guyon and Luxburg {\em et al.}~\cite{guyon2009clustering, von2012clustering}, we evaluate the efficacy of the framework in the context of its end-use. We use an {\em external validation} where ``truth" can be defined a priori over simulated data, enabling a direct evaluation of intrinsic POIs against ``truly interesting'' regions, while ensuring the latent patterns in the generated data mimic the real geospatial dynamics of cars and pedestrians over a region. 

\subsection{Generating synthetic data}
To generate synthetic data for evaluation, we turn to the Simulation of Urban MObility (SUMO) software~\cite{SUMO2012}. SUMO is an open source traffic simulation system capable of generating trajectories of many objects of multiple modalities (e.g. car, truck, person, plane, etc.). Given a shapefile that defines avenues for travel (e.g., a road network, a map of footpaths within a university campus, or the floor plan of a mall or large building), SUMO is able to generate trajectories following the avenues provided. Default parameter settings generate traffic and trajectories in ways that satisfy their measured physical properties have been shown to be incredibly accurate~\cite{SUMO2012}. We use SUMO to generate two simulations of both pedestrian and vehicular traffic under different geographical areas: an  urban region having a mixture of vehicle and pedestrian traffic (the area surrounding The Ohio State University (OSU)), and a suburban area where pedestrian traffic is more prominent (the area surrounding Wright State University (WSU)).
Details about the simulation, the simulation data used in this paper, and the code that produced the resulting evaluation are all publicly available and reproducible online.\footnote{See the following for simulation details: <Anonymized for review purposes.>} The (RSL) cluster tree framework itself is part of a larger open source effort by the author.\footnote{See the following package: <Anonymized for review purposes.>} 

\subsubsection*{Simulation configuration:}
SUMO requires every object to have a {\it trip} defined by departure and destination {\it nodes}, which SUMO refers to as {\it junctions}. Junctions are connected by edges representing a possible travel path. Given a file containing the trip definitions of every object, SUMO dynamically generates {\it routes}, or sequences of edges the object travels along to get from departure junction $A$ to its destination junction $B$. We leave nearly all simulation parameters at their default settings, only modifying simulation length and arrival parameters (binomially distributed arrivals) to generate pedestrian and vehicle demand. 

\begin{figure}
 	\includegraphics[width=0.48\textwidth, height=4cm]{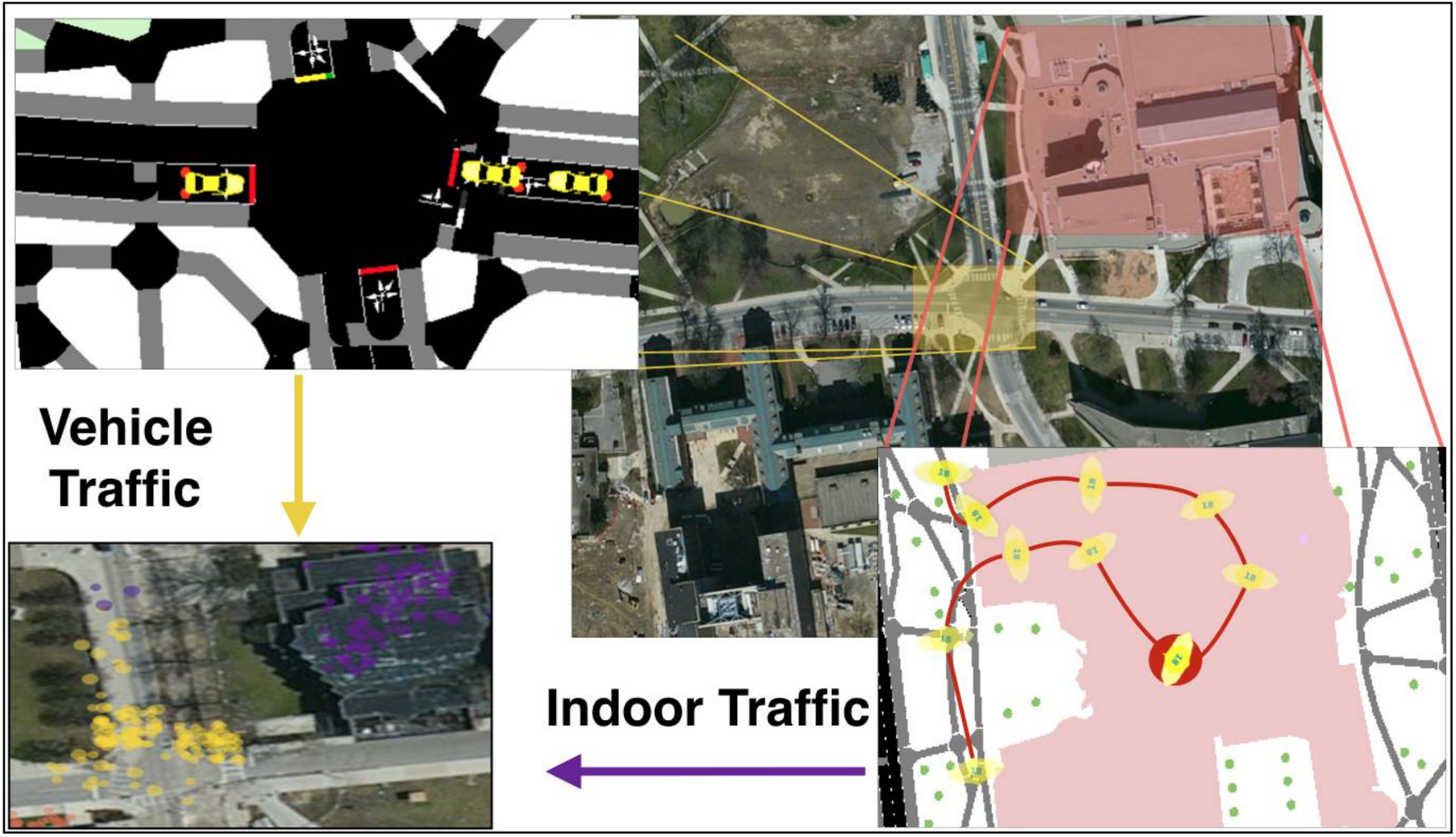}
    \caption{Top-level view of extending SUMO to support indoor pedestrian traffic. 
    Shapefiles defining buildings are loaded into SUMO and registered as junction. 
    If pedestrian-track visits an attached junction during a trip, the simulator chooses
    ordered random set of junctions to follow within the building, exiting after a random period of time.}
    \label{fig:sumo_pic}
\end{figure}

Because pedestrian traffic within unrestricted and indoor areas may constitute intrinsic POIs in a realistic setting, and because SUMO can generate only outdoor pedestrian traffic, we extended SUMO to simulate {\em indoor} pedestrian traffic as well. Figure~\ref{fig:sumo_pic} illustrates how this extension interplays with vehicular traffic generated by SUMO.\footnote{See <Anonymized for review purposes.>} Shapefiles denoting the location of buildings are first loaded into SUMO (the peach colored regions in Figure~\ref{fig:sumo_pic} inlet). Then, within the shapefile, a random number of pedestrian-only junctions are generated within the building and registered to nearby pedestrian-only edges (such as sidewalks). 
If a generated track is labeled as a pedestrian and its trip includes a junction contained within the building region (Figure~\ref{fig:sumo_pic}; lower right inlet), the pedestrian undergoes a random walk {\em within} the junctions generated in the building. This random walk is emulated by choosing a random ordered subset of the generated junctions for a random amount of time. The pedestrian visits these interior junctions and then travels to an `exit junction' attached to the building polygon, continuing along the original (outdoor) route generated by SUMO.
\subsubsection*{Defining truth:} 
Recall that intrinsic POIs are inferred by exemplars representing the specific mobility pattern of interest. With both vehicular and more realistic pedestrian demand generated, the next step in data generation is to define an aggregation function to extract meaningful exemplars. To give a concrete use-case of the proposed framework, we align our experiment with much of the applied literature related to this topic~\cite{zhou07, palma08, zheng2015trajectory} and extract exemplars representing the ``stay points'' of an object. A stay point is a position where objects have stopped or significantly slowed down. Extracting such points from simulated SUMO data is trivial, as the true speed of any traveling object is known at any given time. For pedestrian traffic, we extract trajectory points where pedestrians stopped moving. For vehicular traffic, we extract either a) the points where the vehicles stopped moving or b) the slowest point in a vehicle braking sequence using SUMOs exported braking signals, whichever is available. From these stay points (exemplars), we next establish a mapping between each exemplar and its presence within a ``true" intrinsic POI, allowing external validation. Since exemplars represent object stopped moving, a natural definition of an intrinsic POI is an assignment defined by the mechanism causing such objects to stop. Specifically, we define a building that pedestrians stop within as a ``true" intrinsic POI, as this is very natural and useful grouping.
We follow a similar pattern for vehicular traffic, assigning exemplars a common label if stopped at identical intersections, stop signs or stop light, or other junctions. This mechanistic assignment of creating ``true" intrinsic POIs has the benefit of not only being tractable (in the sense that SUMO provides this information directly), but also being {\em semantically meaningful} in the sense that the mechanisms encouraging objects to stop moving {\it are intrinsic to the geospace}. 

 \subsection{Experimental Design}
\begin{figure*}
\centering
	\renewcommand{\thesubfigure}{a}
	\subcaptionbox{``True'' POIs}{
		\includegraphics[width=0.48\textwidth, height=5.55cm]{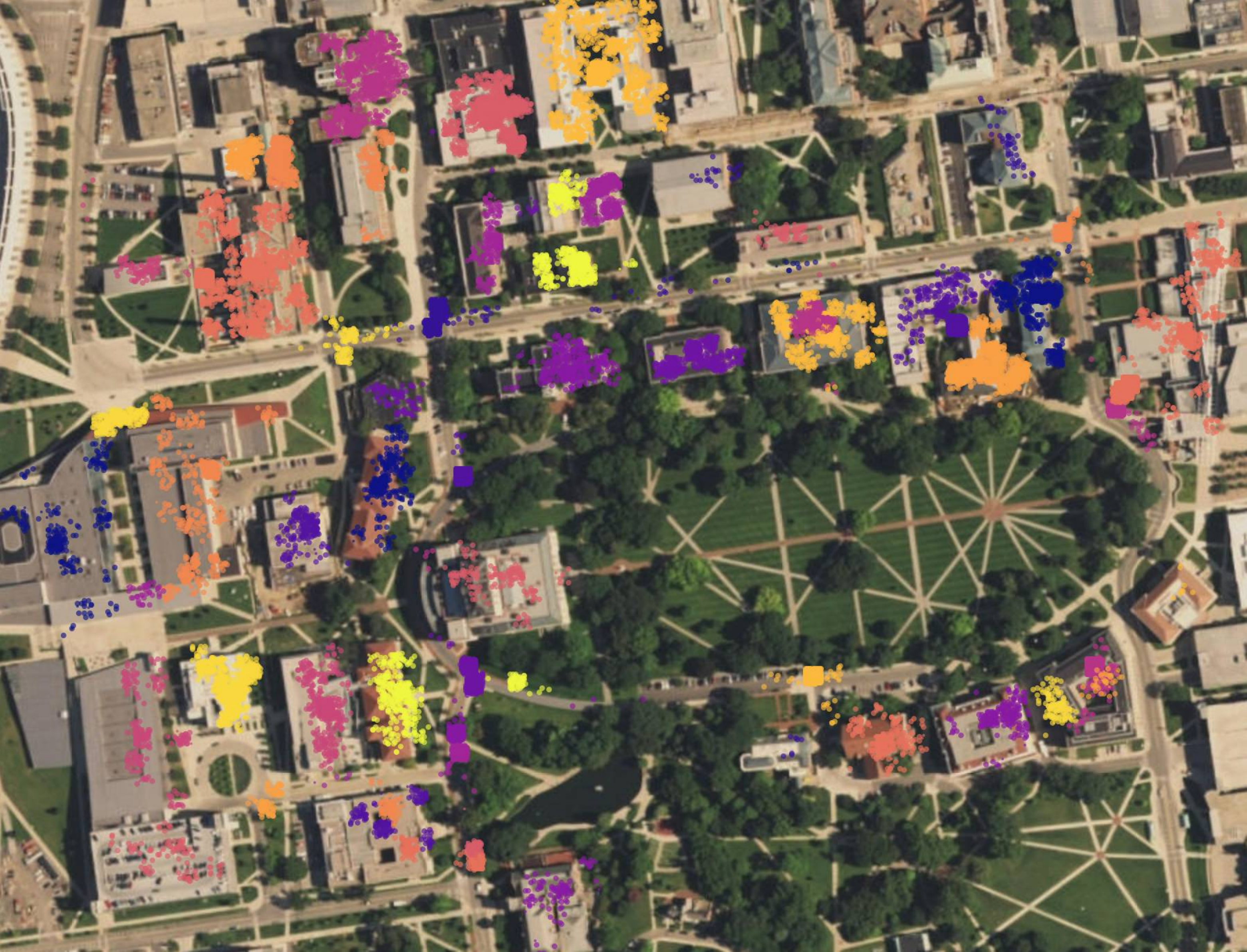}
	}%
	\hfill
	\renewcommand{\thesubfigure}{b}
	\subcaptionbox{Inferred intrinsic POIs}{
		\includegraphics[width=0.48\textwidth, height=5.55cm]{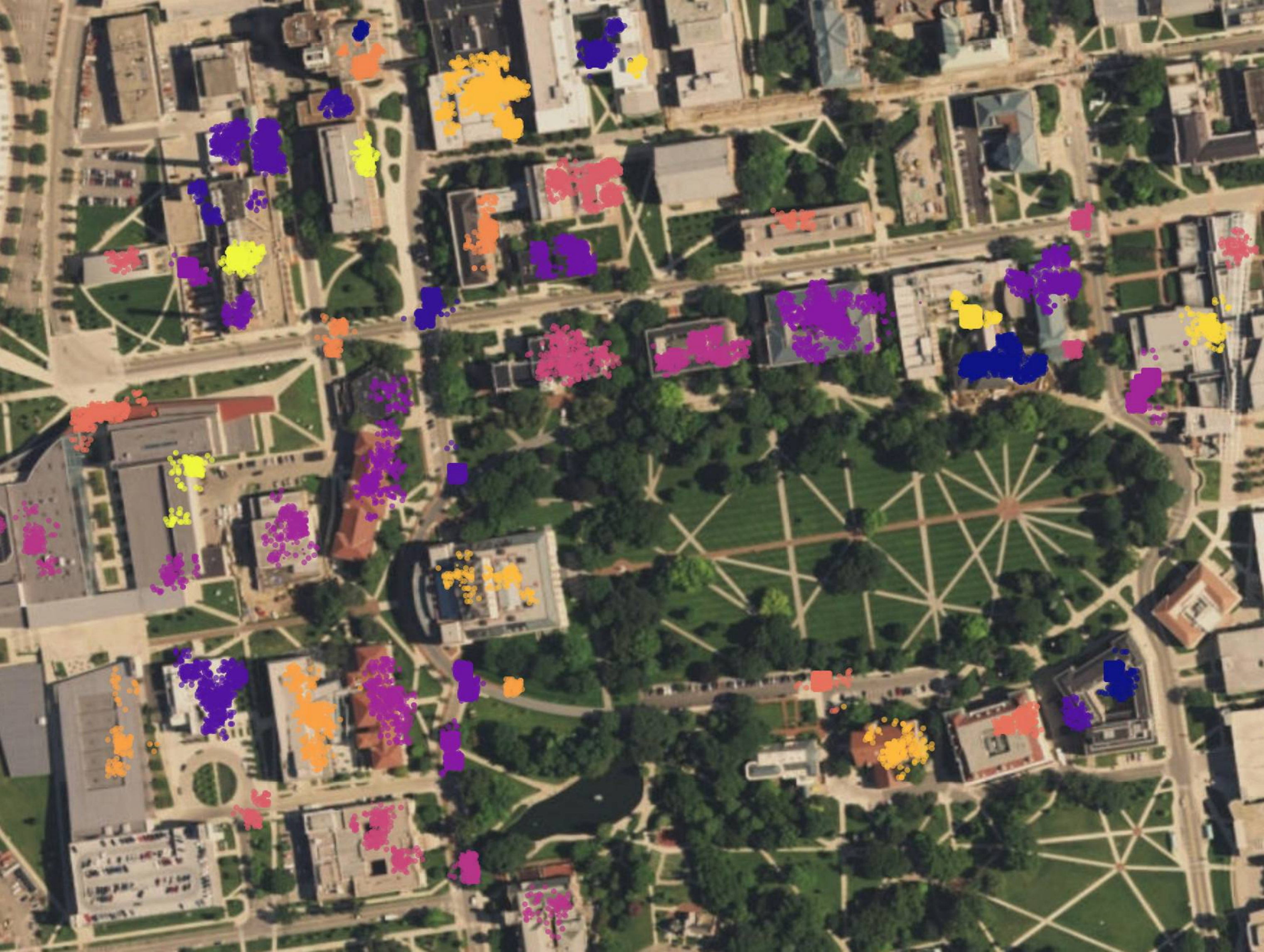}
	}%
	\caption{Intrinsic POI comparison, OSU}
	\label{fig:osu}
\end{figure*}

\begin{figure}
	\centering
	\renewcommand{\thesubfigure}{c}
	\subcaptionbox{``True'' POIs}{
		\includegraphics[width=0.227\textwidth, height=4.85cm]{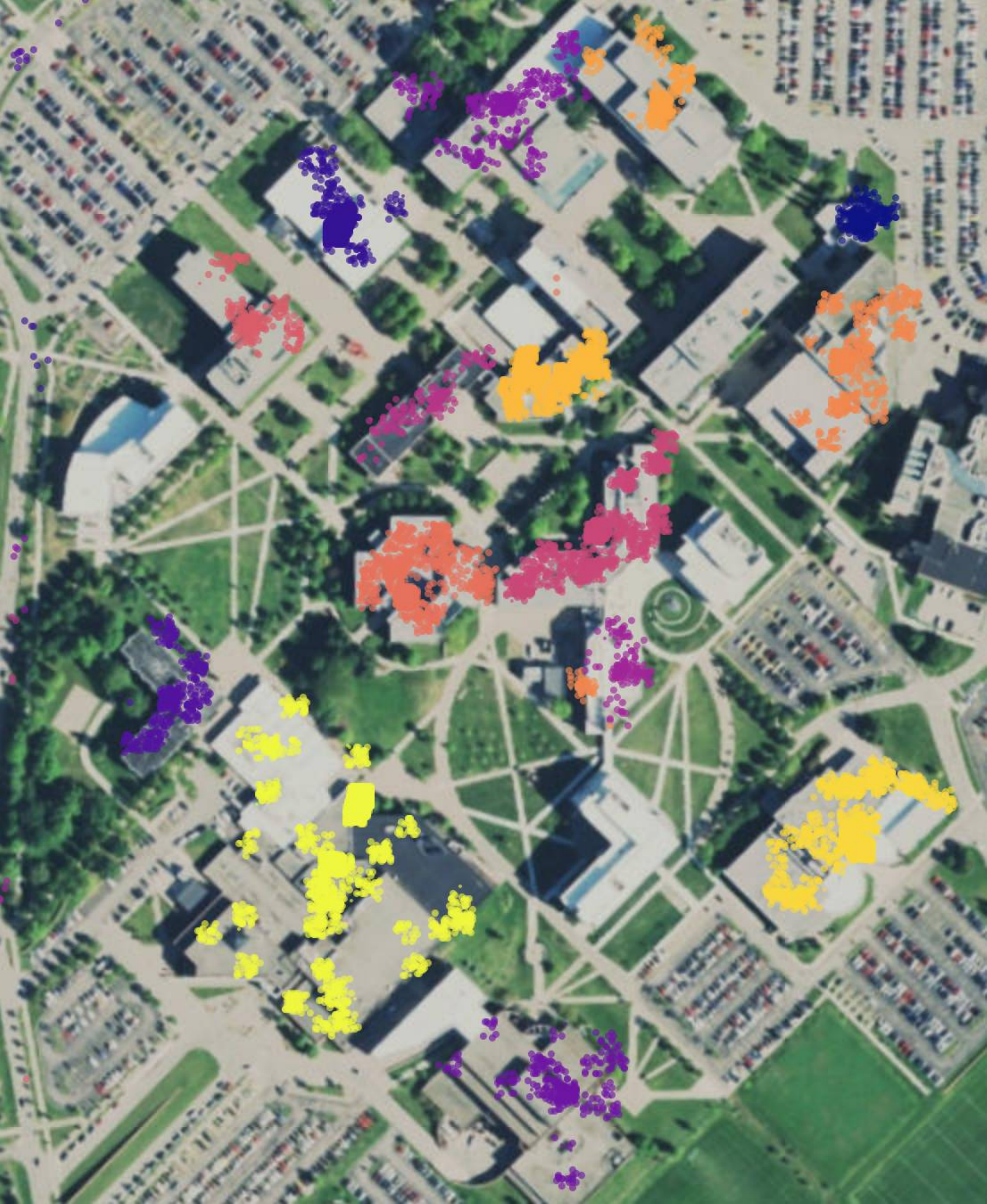}
	}%
	\hfill
	\renewcommand{\thesubfigure}{d}
	\subcaptionbox{Inferred intrinsic POIs}{
		\includegraphics[width=0.227\textwidth, height=4.85cm]{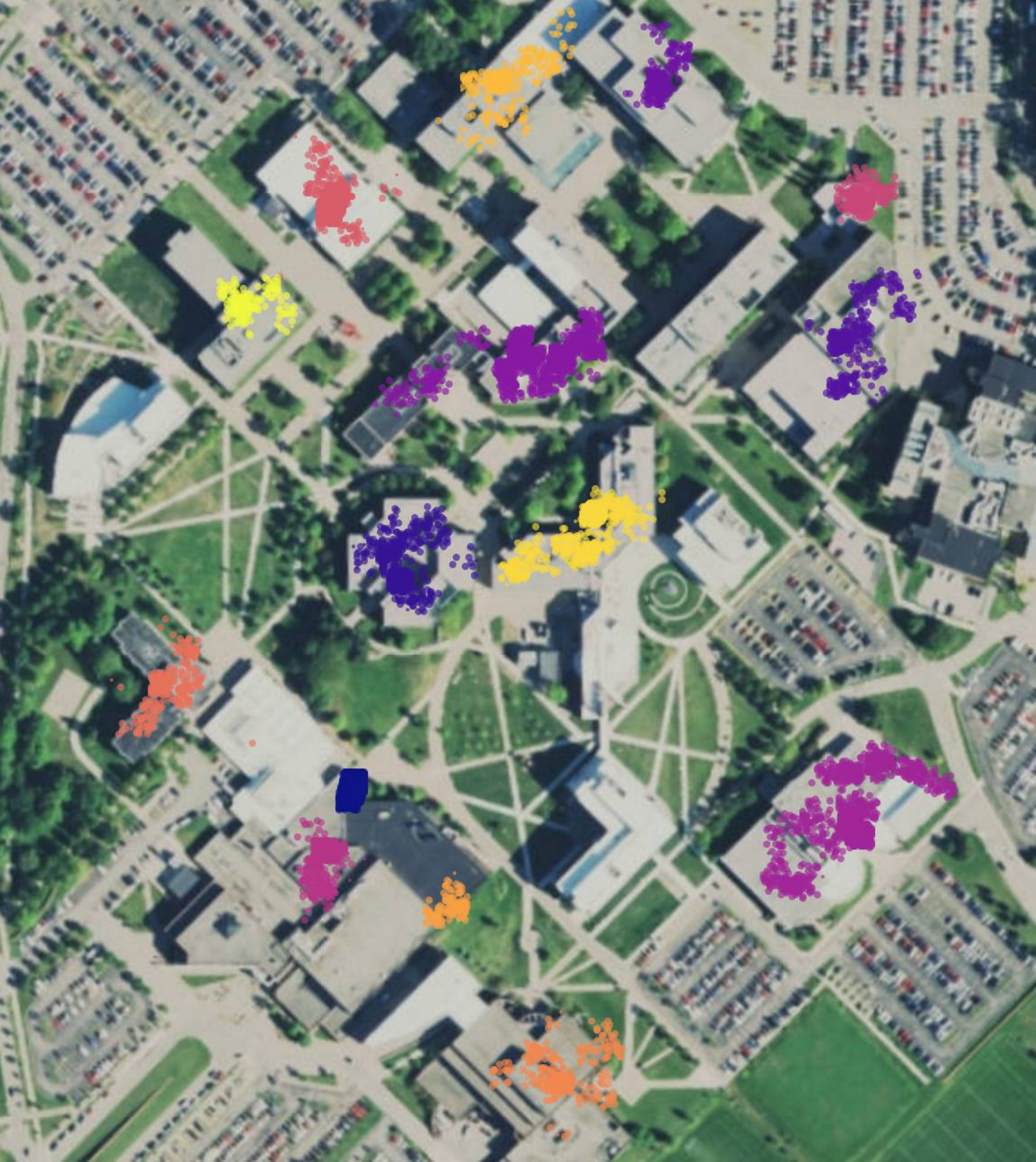}
	}%
	\caption{Intrinsic POI comparison, WSU}
	\label{fig:wsu}
\end{figure}

 To evaluate the fidelity of the POIs extracted by the proposed framework under multiple settings, we run SUMO simulations over the OSU and WSU geospaces with parameter settings reflecting differences between the two regions. These settings are shown in Table~\ref{tab:table_1}. The OSU geospace covers a smaller area, has an equal mix of vehicles and pedestrians, and nearly three times as many buildings. The OSU geospace also has a larger number of roadways and traffic intersections where intrinsic POIs involving vehicles may materialize. Being within the main campus, the WSU geospace has a larger proportion of pedestrian traffic, with few roadways for vehicles to traverse and smaller number of buildings pedestrians may visit. 
 Figures~\ref{fig:osu}(a) and~\ref{fig:wsu}(a) show what this SUMO generated POIs labeling creates for the OSU and WSU campus areas, respectively. 
 Qualitatively, examination of these clusters appear to be reasonable labels of intrinsic POIs. For example, the clusters representing ``true" POIs across OSU in Figure~\ref{fig:osu}(a) finds buildings surrounding the OSU oval quad, particular locations on the ring road around the quad (which tend to be busy OSU intersections for both vehicles and pedestrians), and parking lots around the OSU recreation builds west of the oval to represent POIs. 
 Across WSU in Figure~\ref{fig:wsu}(a), the truth POIs represent each of the major buildings around the campus, with particularly complex, separate areas of movement in WSU's large student union (the yellow points in the large building in the lower left part of the figure). 

\def\tabularxcolumn#1{m{#1}}
\vspace*{-0.085cm}
\begin{table}
\caption{SUMO Simulation Parameters}
	\begin{center}\label{tab:1}
 	\begin{tabularx}{0.48\textwidth} {|| X | X | X | X | X | X | @{}m{0pt}@{} ||} 
 		\hline
 		Region & \# Buildings & \# Veh. & \# Ped. & Region size& Sim. Length  & \\ 
 		\hline\hline
 		OSU & 70 & 2,933 & 2,935 & 342km$^2$  & {\centering 8 hours} & \\ [2ex]
 		\hline
		WSU & 26 & 2,050 & 4,327 & 461km$^2$ & {\centering 6 hours} & \\ [2ex]
 		\hline
 	\end{tabularx}
	\end{center}
	\label{tab:table_1}
\end{table}

%
%

%
%
%
%
%


\subsubsection*{Evaluation Measures:} 
 As discussed at the beginning of this section, the unsupervised nature of intrinsic POI discovery make it difficult to carry out a meaningful evaluation of POI discovery methods using {\em internal} (not requiring `truth' labels) validation measures. 
 Instead, we consider a multifaceted approach: an {\em external}, quantitative evaluation of whether the intrinsic POIs discovered aligns with SUMO generated POIs using the well-known Adjusted Rand Index (ARI)~\cite{hubert1985comparing}, and qualitative evaluation of the quality of the intrinsic POIs our approach unearths as compared to the ``true" intrinsic POIs as defined above. 
 The Rand-family of indices were chosen due to their transparency and simplicity---although, whereas the traditional RI measures the proportion of pairwise agreements between two partitions, the ARI also adjusts the score based on the expected value of agreements under the null hypothesis that the agreements were completely random, and thus is what we report. 

\subsubsection*{Algorithms Compared:} We further compare the fidelity of the intrinsic POIs extracted by the proposed framework against other clustering algorithms commonly used for POI discovery from trajectories. We either downloaded the implementation of, or implemented ourselves, a number of these algorithms for comparison. Aside from RSL, the selected methods includes the well-known density-based algorithms DBSCAN~\cite{ester1996density} and OPTICS~\cite{ankerst1999optics}, the widespread hierarchical algorithms single linkage (SL), average linkage (AL), and wards criterion (WL)~\cite{murtagh2014ward}, along with the partitioning-like algorithms $k$-means and CLARA~\cite{clara}. These algorithms were chosen due to their relevance to this problem, wide-spread availability, known success in the clustering world. 

\subsubsection*{Parameter Settings:} 
\label{sec:params} Clustering algorithms generally require parameter-tuning in order to `fit' to a given data set, but the number and semantics of these parameters often changes with the algorithm used, leaving comparisons between parameter settings difficult. 
Although most hierarchical algorithms carry no free parameters to create a (hierarchical) {\em set of solutions}, they do require either a threshold value ($h$) or the exact number of clusters to extract ($k$) to be specified to extract a `flat' clustering. Similarly, $k$-means and CLARA also require $k$ to be specified a priori. Because the $k$ parameter has the same interpretation in multiple algorithms, we will use $k$ to refer to {\em the number of clusters extracted}. Density based algorithms have multiple parameters with interpretations compared to the aforementioned algorithms. For example, DBSCAN requires a {\it minimum} cluster size parameter $\mathit{minPts}$ and a distance (or `scale') threshold $\epsilon$ to be set. OPTICS, often cited as  extension to DBSCAN, is an ordering algorithm that---given a parameter setting for $\mathit{minPts}$---can be be used to extract either a flat, DBSCAN-like cluster extraction or a simplified hierarchy using a either the a distance threshold $\epsilon'$ or a reachability-based threshold $\xi$, respectively. The DBSCAN-like cluster extraction is reported here. RSL requires the setting of $\alpha$ and $k$, the former relating to scaling the connection radii used to connect components, and the latter to the saliency of cluster estimates. Note that in RSL, $k$ is more similar to the $minPts$ parameter in that it is a {\em minimum} neighborhood parameter. The number of clusters is automatically determined by optimized the defined relative excess of mass functional from Section \ref{sec:methodology}. 

Each algorithm reflects a large set of possible solutions over its parameter setting. Choosing a single parameter setting for evaluation would represent a source of possible {\em bias}. Rather, we employ a more comprehensive approach by comparing a wide range of parameter settings for each algorithm. To define these ranges, let $\texttt{seq}(x,y,s)$ denote the sequential range operator, skipping $s$ values in the sequence of integers from $x$ to $y$. For example, $\texttt{seq}(1,n,1) = \{1, 2, \dots, n\}$, and  $\texttt{seq}(1,n,i) = \{1, 1 + i, \dots, n - i, n\}$. 
For the hierarchical clustering algorithms (SL, AL, and WL) the number of flat clusters extracted $k$ is varied in the range $\texttt{seq}(2,n_t,1)$ where $n_t$ is the number of ``true" POIs assigned by SUMO. We see this as a reasonable strategy, as it gives a better view of how multiple levels extracted from the hierarchy matched the data set as well as how well the merge criterion (or linkage function) collectively captures the true POIs in the geospace. We use the same range to vary $k$ for the $k$-means and CLARA algorithms. The density based methods DBSCAN and OPTICS are evaluated by first varying $\mathit{minPts}$, and then (for each value of $minPts$) by varying the scale parameters $\epsilon$ and $\epsilon'$, respectively. Recall $\mathit{minPts}$ relates to a {\it minimum} neighborhood value that constitutes a cluster. Thus, and to allow the testing to be tractable, we set $\mathit{minPts}$ to reflect the possible sizes of the POIs, along the {\em quantiles} $q_{n_t} = \texttt{seq}(0.10,0.95,0.025)$ corresponding to the the number of exemplars per POI in the SUMO ``truth'' data. The distance thresholds $\epsilon$ for DBSCAN and $\epsilon'$ OPTICS are also varied along the quantiles $\texttt{seq}(0.01,0.20,0.01)$ of the pairwise distances computed over the data set. Since all density-based methods mark points that fall in areas of the data set not sufficiently dense as `noise' according to a scale parameter---leading to severe overfitting if not guided with a measure like stability---all density-based solutions were deemed only valid if at least $75\%$ of the data is classified with a non-noise label. Finally, the RSL also contains two parameters, a $k$ value, and an $\alpha$ parameter. We use Chaudhuri {\em et. al}'s analysis to determine how to set these. RSL was shown to have optimal rates of convergence when $\alpha \geq \sqrt{2}$, so we leave it at that constant value ($\sqrt{2}$). Similarly, the rate only holds for $k$ at least as large as $d \log(n)$, where $d$ is the dimensionality of the data set ($d=2$ in this case). After varying through the small set of $k$ values in a similar fashion as was performed for DBSCAN and OPTICS ($k \in q_{n_t} \forall k >= d \log(n)$). 
In total, 2,196 and 1,995 cluster configurations were performed for the OSU and WSU simulations respectively, totalling 4,191 reported configurations.  

 \subsection{Validation Testing and Discussion}
 \begin{figure*}\label{fig:res}
\begin{multicols}{2}
    \includegraphics[width=\linewidth, keepaspectratio]{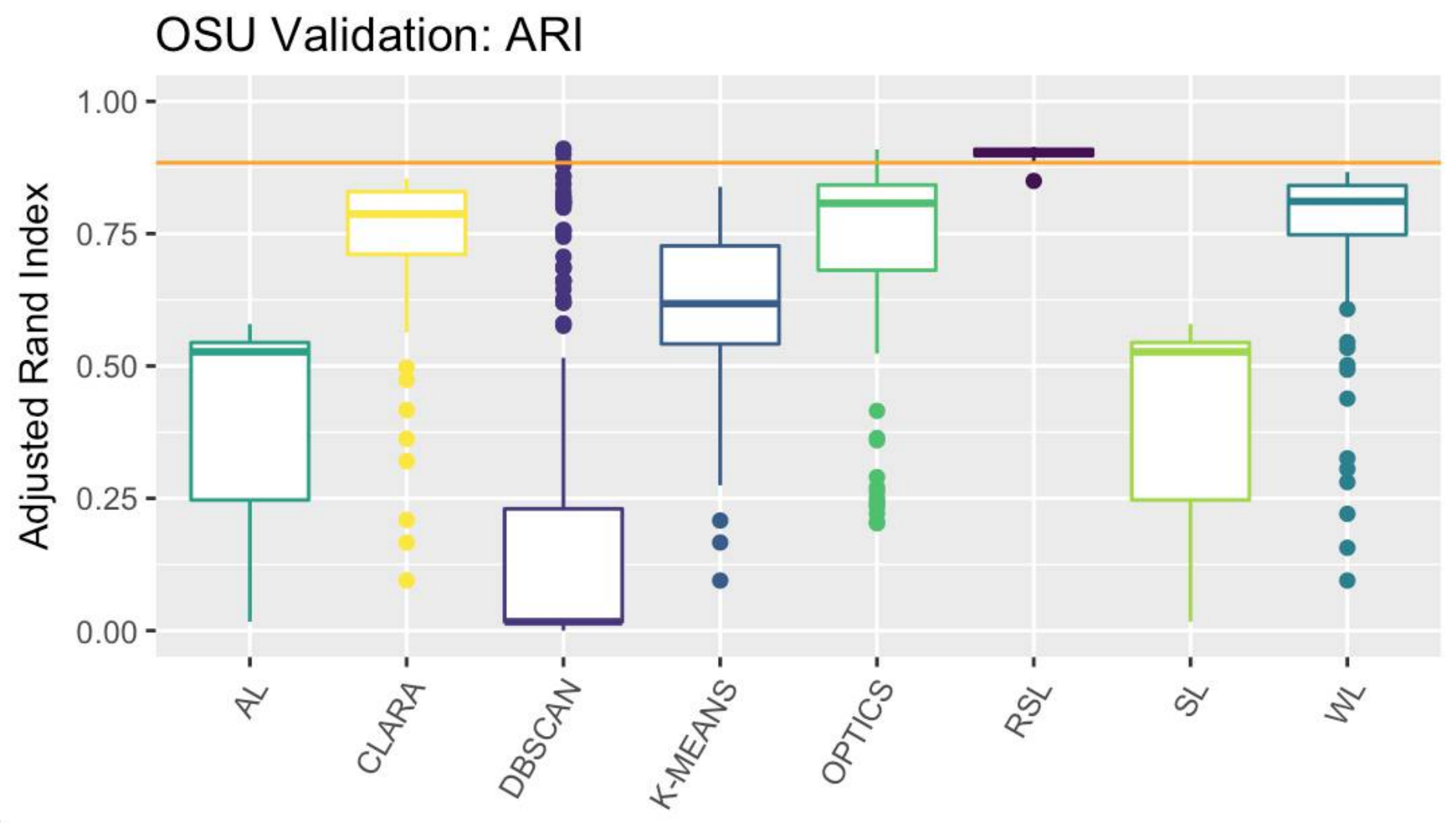}\par 
    \includegraphics[width=\linewidth, keepaspectratio]{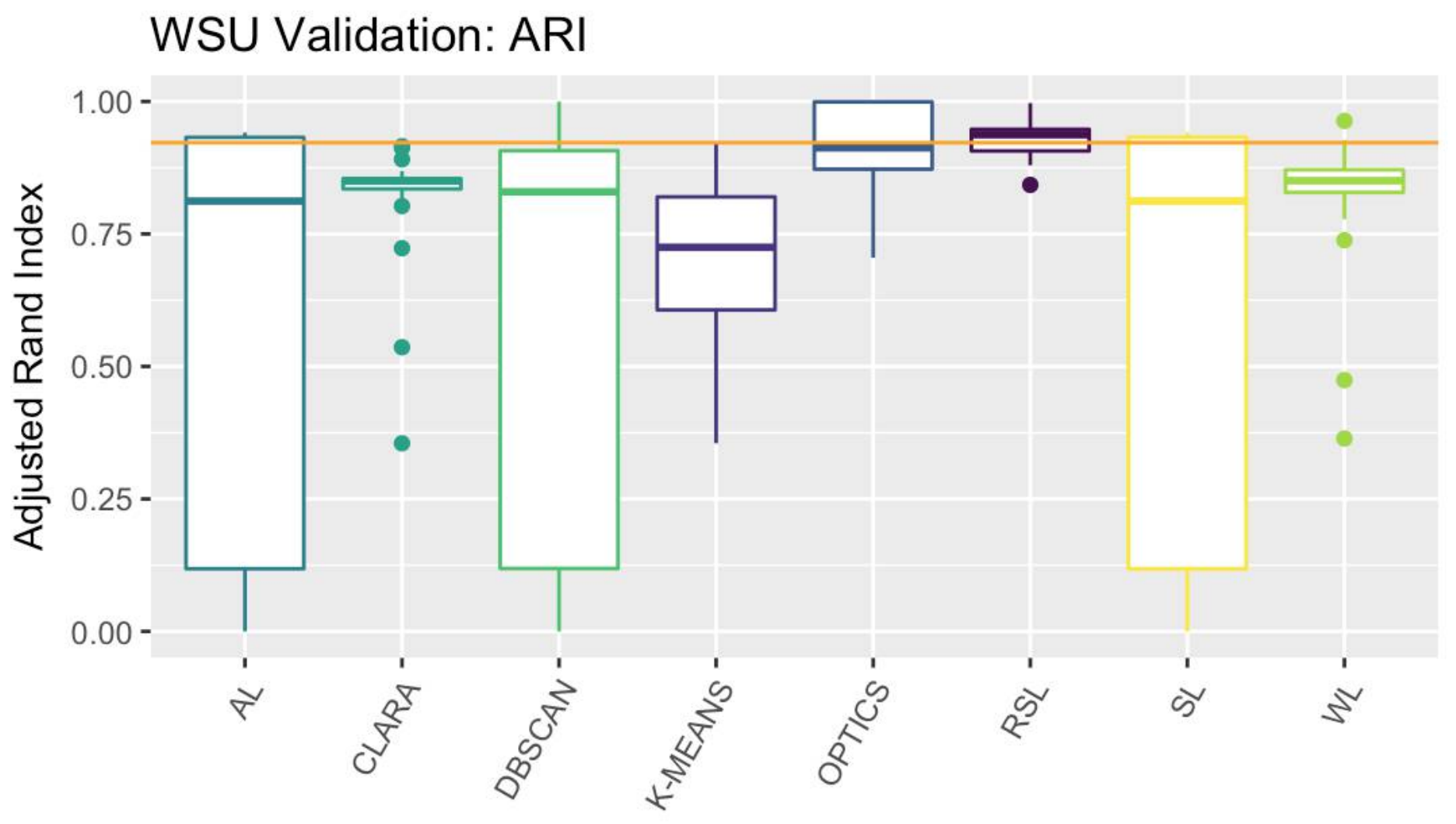}\par 
 \end{multicols}
 \vspace*{-0.475cm}
\caption{The distribution of Adjusted Rand Index (ARI) scores of various clustering algorithms (after varying free parameters). The orange line corresponds to the ARI of the proposed framework.}
\label{fig:res}
\end{figure*}

\subsubsection*{Qualitative comparison to truth:}
Figures~\ref{fig:osu} and~\ref{fig:wsu} compare the intrinsic POIs discovered by our framework
agains the simulation's ``true" POIs. Recall that points with low density are discarded as noise (not shown). Direct comparison of the true POIs defined by the simulation show clear similarities. Over the OSU simulation in Figure~\ref{fig:osu}(b), the framework recovers intrinsic POIs within buildings no matter its shape, the density, or closeness to other buildings. It also recovers intrinsic POIs over parking lots and street intersections around the OSU oval. Some buildings are decomposed into a collection of individual intrinsic POIs. For example, the eastern-most large campus building by the northeast corner of the oval contains three separate intrinsic POIs: one at its entrance by the road, another in the center of the building, and a third at its back entrance. Although these labels may not match what SUMO assigned, they are in some sense more natural, i.e. it's quite possible for large buildings to have dense, isolated areas of people movement.

Looking at the intrinsic POIs over the WSU dataset in Figure~\ref{fig:wsu}(b), we find each building in general is recovered as an intrinsic POI and align in shape compared to the shape of the ``true" POIs from Figure~\ref{fig:wsu}(a). We also note that the framework determines that some movement within buildings but covering very small areas were found not be significant enough be an intrinsic POI. Large buildings also showed further decomposition like the OSU simulation. For example, in the WSU student union (large building in the lower left corner of the figure) the framework defines intrinsic POI's at the center and two back exits from the building. 

\subsubsection*{Quantitative comparison to other approaches:}
The proposed intrinsic POI framework measured ARI scores of $0.966$ on the OSU data set and $0.922$ for the WSU data set. Note that this is not the maximum ARI of any RSL solution, but the ARI of the solution found using the highest predefined notion of stability, determined completely without {\em any} knowledge of the surrounding geographical area.  RSL performed consistently in terms of having low variability compared to other algorithm, with overall high similarity to the semantically driven SUMO assigned locations. Figure~\ref{fig:res} shows the distribution of the ARI for the algorithms we compared our method against with the parameter settings discussed in Section~\ref{sec:params}. The orange line corresponds with the ARI of the proposed framework, which compares favorably with best possible settings of other algorithms. Note that although DBSCAN (like others) performed well with {\it very specific} configurations of $\mathit{minPts}$ and $\eps$, the settings of these parameters is often not very intuitive in unsupervised scenarios where the truth is unknown (and thus external measures like ARI cannot be computed). Of the hierarchical algorithms, we see the impact of the linkage criterion used and how they are influenced by the `shape' of the true clusters. For example, SL clustering performed fairly well on the well separated WSU data set, but substantially lower on the more density-varied OSU data set. This is reflected in AL as well. $k$-means was able capture much of the true clustering structure with the right parameter settings for the WSU simulation (with max/mean ARI scores of $(0.92, 0.71)$), however exhibited degraded performance when the POIs were less separated  in OSU data set ($(0.61, 0.84)$). OPTICS, with a few specific parameter settings, performed well on the WSU data set, however again suffered on the more variable-scale OSU data set.

\section{Related Research}\label{sec:related_research}
The trajectory field has been largely progressed by ``extensive and intensive individual efforts''~\cite{zheng2015trajectory}. Nonetheless, conceptual models have been proposed for how to deal with the patterns within trajectories and how to relate such patterns to geographical areas of interest for various purposes. One such model postulates that trajectories and their spatiotemporal patterns are essentially driven by the semantics the application associates with trajectory itself~\cite{parent2006conceptual}~\cite{spaccapietra2008conceptual}, and have contributed significantly to the ``Stops and Moves of Trajectories'' (SMoT) family of classification algorithms~\cite{alvares2007model, palma08, rocha2010db}, where the premise of the analysis is that by partitioning trajectory data into a labeled set of `stop' and `move' segments, one can take then annotate these segments with semantic information, derive specific mobility patterns, and as a result {\it discover} `interesting' locations. 
Alvares {\em et al.} developed IB-SMoT to find interesting positions based on semantic annotations describing the places a trajectory visited~\cite{alvares2007model}. Palma {\em et al.} reduced IB-SMoT's reliance on prior knowledge about positions that are likely to be interesting by incorporating the speed at which tracks are traveling with their variation CB-SMoT~\cite{palma08}. DB-SMoT finds clusters of common trajectories based on similar direction changes and stopping points~\cite{rocha2010db}. Zhou {\em et al.} tackle the problem of finding positions of interest to an individual track based on data about a track's location preferences, position over time, and tags of locations provided by web services such as Google Maps~\cite{zhou07}. 

Many related efforts encode or are reliant on varying notions of an ``interesting place'' using, for example, techniques from natural language processing (NLP)~\cite{gabrielli2014tweets}, data clustering~\cite{rocha2010db, zhou07, uddin2011finding}, sequential pattern mining~\cite{xiao2014inferring}, and social network analysis~\cite{zheng2011location} methods. Zheng {\em et. al} pioneered the use of `stay points', corresponding to an aggregation of consecutive GPS points that collectively are within a user-supplied time and distance threshold, thereby characterizing a `virtual' location~\cite{zheng09, zheng2010geolife}. It is interesting to note that Zheng {\em et. al} used OPTICS to create a hierarchical clustering `stay points' for an LBS-type application with Microsoft called `Geolife'~\cite{zheng2010geolife}. Indeed, Zheng anticipated a number of developments in the trajectory mining field~\cite{zheng2015trajectory}; the theoretical cluster tree may be viewed as a more statistically based conception of the`Tree Based Hierarchical Graph' that is used to represent POIs in that application as well. 

It's worth noting that our definition of an intrinsic POI, having a more theoretical foundation in density-based clustering, is both conceptually very similar to OPTICS and computationally, more recently, to Hierarchical DBSCAN (HDBSCAN)~\cite{campello2013density}. There exist a number of commonalities between both OPTICS and DBSCAN and the theory of the cluster tree. A comprehensive exposition of this relationship is beyond the scope of this paper, see Campello {\it et. al}~\cite{campello2015hierarchical} for a thorough review of the subject. Although it's not mentioned in such efforts, the usage of RSL with a relative excess of mass functional to cluster extraction is equivalent to the flat clusters ``HDBSCAN'' extracts with a setting of $\alpha = 1$ and $k = \mathit{minPts}$. However, the asymptotic consistency of the setting of the pair $(\alpha = 1, k \sim d \log n)$ has not been established~\cite{chaudhuri2010rates}. When $alpha = 1$, $k$ must be much larger, exponential in the dimensionality of the data set, $d$. Thus, we use RSL with $\alpha \geq \sqrt{2}$. 

\section{Concluding Remarks}\label{sec:concluding_remarks} 
This paper proposed a general framework for intrinsic POI discovery, without needing to rely on external gazetteers, based on recent theoretical advances in hierarchical, nearest neighbor density estimation. It discussed a conceptually sound basis for automated POI discovery specifically in the context of geospatial data, and introduced a framework that provides a rigorous and usable solution to an applied domain primarily dominated by intuitively reasonable, but heuristically-based methods. With novel extensions to SUMO to support pedestrian movement in buildings, an evaluation of simulated trajectory data over diverse geographical areas supports the conclusion that the proposed framework is a useful tool for extracting intrinsic POIs. The framework has both theoretical guarantees and practical benefits, requires no ad hoc parameter tuning, and exhibits improved fidelity against common approaches over thousands of parameter settings. 

In future work, with the help of the asymptotic analysis done by Chaudhuri {\em et. al}, we plan to develop model-selection techniques for POI extraction. This is imperative in exploratory settings, such as large urban environments where the number of POIs is not known ahead of time, there is little useful knowledge to gain from ad hoc or heuristic-based cluster analysis, especially when the solution space is large. By relating the concept of a POI to the theory of the cluster tree, RSL and associated estimators enable future theoretical work may further augment models reliant on POI data, such as location recommendation systems, collaborative filtering techniques, or social networking models built from POI data, such the the `Location-Based Social Networks' reviewed in ~\cite{zheng2011location, zheng2015trajectory}. 


\balance
\setcitestyle{numbers,sort&compress}
\bibliographystyle{ACM-Reference-Format} 
\bibliography{poi}
\end{document}